# Visual Detection of Personal Protective Equipment and Safety Gear on Industry Workers

Jonathan Karlsson[1], Fredrik Strand[1], Josef Bigun[1][a], Fernando Alonso-Fernandez[1][b]
Kevin Hernandez-Diaz[1][c], Felix Nilsson[2]
[1]*School of Information Technology (ITE), Halmstad University, Sweden*
[2]*HMS Industrial Networks AB, Halmstad, Sweden*
jjonathankarlsson@gmail.com, fredrik00.strand@gmail.com{josefbigun, feralo, kevher}@hh.se, fenil@hms.se

Abstract: Workplace injuries are common in today's society due to a lack of adequately worn safety equipment. A system that only admits appropriately equipped personnel can be created to improve working conditions. The goal is thus to develop a system that will improve workers' safety using a camera that will detect the usage of Personal Protective Equipment (PPE). To this end, we collected and labeled appropriate data from several public sources, which have been used to train and evaluate several models based on the popular YOLOv4 object detector. Our focus, driven by a collaborating industrial partner, is to implement our system into an entry control point where workers must present themselves to obtain access to a restricted area. Combined with facial identity recognition, the system would ensure that only authorized people wearing appropriate equipment are granted access. A novelty of this work is that we increase the number of classes to five objects (hardhat, safety vest, safety gloves, safety glasses, and hearing protection), whereas most existing works only focus on one or two classes, usually hardhats or vests. The AI model developed provides good detection accuracy at a distance of 3 and 5 meters in the collaborative environment where we aim at operating (mAP of 99/89%, respectively). The small size of some objects or the potential occlusion by body parts have been identified as potential factors that are detrimental to accuracy, which we have counteracted via data augmentation and cropping of the body before applying PPE detection.

## 1 INTRODUCTION

A study made by the International Labour Office concluded that approximately 350,000 workers die annually due to workplace accidents. In addition, there are about 270 million accidents at work leading to absence for three or more days. A high level of productivity is associated with good occupational safety, and health management (ILO, 2005).

The U.S. Bureau of Labor Statistics shows statistics on injury conditions for various industries. The report concludes that construction has the most considerable fatal injuries. Statistics show that 5,039 American construction workers died between 2016 and 2020 due to a deficiency of personal protective equipment, safety deficiencies, and other accidents. This is a high number compared to the occupations like (BLS, 2021): agriculture, forestry, fishing, and hunting (2832); mining, quarrying, and oil and gas extraction (536); or finance and insurance (129). The National Institute for Occupational Safety and Health (NIOSH) has concluded that 25% of all construction fatalities are caused by traumatic brain injury (TBI) (Nath et al., 2020), and 84% of these fatal injuries are due to not wearing hardhats (Ahmed Al Daghan et al., 2021). Heavy machinery and trucks are standard within construction sites, and it is essential to ensure high visibility of the workers to prevent poten-tially fatal injuries (Ahmed Al Daghan et al., 2021). Many fatal deaths could potentially be avoided with proper protective equipment, such as vests with reflexes increasing visibility.

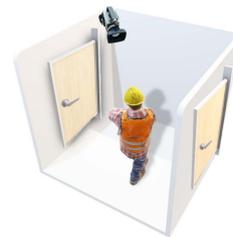

Figure 1: Airlock room to separate controlled environments.

[a] https://orcid.org/0000-0002-4929-1262
[b] https://orcid.org/0000-0002-1400-346X
[c] https://orcid.org/0000-0002-9696-7843

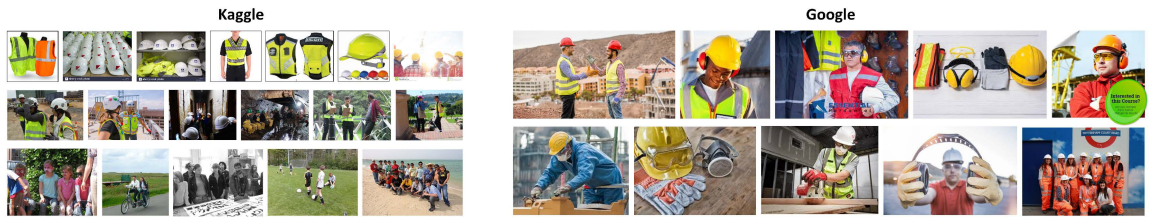

Figure 2: Left: Examples of images from the Kaggle dataset of hardhats and safety vests with positive examples (rows 1,2) and negative examples (row 3). Right: Examples of images gathered from Google to complement the safety gloves, safety glasses, and hearing protection classes. See Section 3.1 for further details.

AI-driven applications are increasingly utilized in several environments where security and protection are prioritized. This work aims to develop a system that helps enforce Personal Protective Equipment (PPE) usage. PPE detection could also be combined with facial recognition to determine whether people are authorized to access an area of the production floor, which is the aim of the project that motivates this research. In the present paper, however, we have only focused on developing and evaluating PPE detection compliance. Based on the literature review of the next section, PPE is an occurring topic in object detection and is often used as a monitoring system. However, the combination of face recognition and PPE detection is not very common. Also, the number of PPE classes in existing works is usually one or two (normally hardhat or vest), but we aim at detecting five different objects (hardhat, safety vest, safety gloves, safety glasses, and hearing protection).

The contribution of this work is thus to increase the number of target classes to five and, in a later stage, combine facial recognition and PPE object detection. Fast execution and response are important for a user-friendly system. Therefore, another goal is to have real-time detection. Combining object detection with identity recognition via facial analysis can produce virtual key cards that only allow authorized staff with the proper gear to enter a restricted area via, e.g., an airlock room (Figure 1). An airlock room is a transitional space with two interlocked doors to separate a controlled area, where different checks can be made before granting access. A system like this could save thousands of lives annually and prevent severe injuries with object detection and facial recognition tools. We thus aim ultimately at increasing the safety of industry workers in hazardous environments.

This project is a collaboration of Halmstad University with HMS Networks AB in Halmstad. HMS develops and markets products and solutions within ICT for application in several industrial sectors (HMS, 2022). They explore emerging technologies, and one crucial technology is AI, where they want to examine and showcase different possibilities and applications of AI and vision technologies, e.g. (Nilsson et al., 2020), which may be part of future products.

## 2 RELATED WORKS

Several projects (Nath et al., 2020; Jana et al., 2018; Protik et al., 2021) use machine learning or deep learning to detect PPE in different conditions. Object detection and facial recognition are also common topics within machine learning.

In (Protik et al., 2021), the objective is to detect people wearing face masks, face shields, and gloves in real time. They used the YOLOv4 detector (Bochkovskiy et al., 2020) with a self-captured database of 1,392 images gathered from Kaggle and Google, suggesting that real-time operation is feasible (a score of 27 frames per second (FPS) was achieved). The achieved mean average precision (mAP) is 79%. The authors also implemented a counter function for objects and record keeping.

In (Ahmed Al Daghan et al., 2021), the authors developed a system to notify a supervisor when a worker is not wearing the required PPE (helmet and vest). It was achieved through object detection with YOLOv4, and facial recognition with OpenFace (Baltrušaitis et al., 2016) that identified the worker and its gear via CCTV in the construction site. An alert system notifies the supervisor via SMS and via email with an attached image of the incident. Supervisors are also informed if non-authorized personnel enter the construction site. They used 3,500 images collected from Kaggle, with 2,800 to train the model during 8,000 iterations. The attained mAP is 95% (vest), 97% (no vest), 96% (helmet), 92% (no helmet).

Three different approaches are proposed in (Nath et al., 2020) based on YOLOv3 to verify PPE compliance of hard hats and vests. The first approach detects a worker, hat, and vest separately, and then a classifier predicts if the equipment is worn correctly. The second approach detects workers and PPE simultaneously with a single CNN framework. The last approach first detects the worker and uses the cropped image as input to a classifier that detects the presence of PPE. The paper uses ∼1,500 self-obtained and annotated images to train the three approaches. The second approach achieved the best performance, with a mAP of 72.3% and 11 FPS on a laptop. The first approach, although having lower mAP (63.1%), attained 13 FPS.

The paper (Fang et al., 2018) proposed a system to detect non-hard hat use with deep learning from a far distance. The exact range was not specified, just small, medium, and large are mentioned. Faster R-CNN (Ren et al., 2015) was used due to its ability to detect small objects. More than 100,000 images were taken from numerous surveillance recordings at 25 different construction sites. Out of these, 81,000 images were randomly selected for training. Object detection is outdoors, and the images were captured at different periods, distances, weather, and postures. The precision and recall in the majority of conditions are above 92%. At a short distance, like the one that we will employ in this work, they achieved a precision of 98.4% and a recall of 95.9%, although the system only checks one class (helmet).

In (Zhafran et al., 2019), Faster R-CNN is also used to detect helmets, masks, vests, and gloves. The dataset consists of 14,512 images, where 13,000 were used to train the model. The detection accuracy was measured at various distances and under lighting but with a distracting background. Helmet resulted in 100% accuracy, masks 79%, vest 73%, and gloves 68% from a 1-meter distance. Longer distances contributed to poorer results, with detection at a 5-meter distance or above deemed unfeasible for all classes. Reduced light intensity was also observed to have a high impact on the smaller objects (masks, gloves).

The paper (Delhi et al., 2020) developed a framework to detect the use of PPE on construction workers using YOLOv3. They suggested a system where two objects (hat, jacket) are detected and divided into four categories: NOT SAFE, SAFE, NoHardHat, and NoJacket. The dataset contains 2,509 images from video recordings of construction sites. 1. The model reported an F1 score of 0.96. The paper also developed an alarm system to bring awareness whenever a non-safe category was detected.

In (Wójcik et al., 2021), hard hat wearing detection is carried out based on the separate detection of people and hard hats, coupled with person head keypoint localization. Three different models were tested, ResNet-50, ResNet-101, and ResNeXt-101. A public dataset of 7,035 images was used. ResNeXt-101 achieved the best average precision (AP) at 71% for persons with hardhat detection, but the AP for persons without hard had was 64.1%.

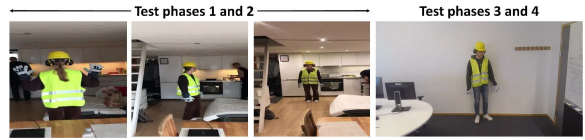

Figure 3: Examples of images from volunteers to be used as test images of the evaluation phases (see Section 3.3).

# 3 METHODOLOGY

## 3.1 Data Acquisition

The five object classes to be detected in this work include hardhat, safety vest, safety gloves, safety glasses, and hearing protection. Two data sources have been used to gather data to train our system. A dataset from kaggle.com[1] (Figure 2, left) with labeled hardhats and safety vests was utilized, containing 3,541 images, where 2,186 were positive, i.e. with hardhats, safety vests, or both present. Negative images are pictures of people in different activities indoors and outdoors. An additional 2,700 images were collected and annotated from Google Images with different keyword searches to complement the remaining classes (Figure 2, right), given the impossibility of finding a dataset containing all target classes. Preliminary tests showed that, for example, caps could be predicted as hardhats. This was solved by adding images of caps as negative samples of the training set to help the model to learn the difference. The available data has been augmented with different modifications (saturation, brightness, contrast, blur, and sharpness) to increase the amount of training data of some underrepresented classes, as it will be explained during the presentation of results in Section 3.3.

## 3.2 System Overview

Several algorithms exist for object detection, e.g.: RetinaNet (Lin et al., 2020), Single-Shot MultiBox Detector (SSD) (Liu et al., 2016), and YOLOv3 (Redmon and Farhadi, 2018). Comparatively, RetinaNet usually has a better mean average precision (mAP), but it is too slow for real-time applications (Tan et al., 2021). SSD has a mAP between RetinaNet and YOLOv3. YOLOv3 has the worst mAP but is faster and suits real-time applications. In addition, it is the fastest to train, so one can adapt quickly to changes in datasets. YOLOv4 (Bochkovskiy et al., 2020) is an improvement of YOLOv3, with an increase in mAP and frames per second of 10% and 12%, respectively. YOLOv4 was also the newest YOLO version when this research was conducted, making it the choice of object detection algorithm for this work.

---

[1] https://www.kaggle.com/datasets/johnsyin97/hardhat-and-safety-vest-image-for-object-detection

Table 1: Data variability conditions in the different test phases

| test phase | training data | test data | body crop |
|---|---|---|---|
| 1 | Kaggle + Google | 3 people (two males, one female), each person wears all PPEs<br>119 images (3 meters), 127 images (5 meters), 111 images (7 meters)<br>High variability: different angles, light, and background | no |
| 2 | Kaggle + Google augmented | Same as phase 1 | no |
| 3 | Kaggle + Google augmented | 2 people (different than above), each person wears all PPEs<br>100 images (at 3 meters), 100 images (at 5 meters)<br>Low variability: camera at head level, static, frontal, controlled background | yes |
| 4 | Kaggle + Google augmented<br>2 people (of phases 1-2) + augmented | same as phase 3 | yes |

Table 2: Results of the first test phase.

| Object | Precision/Recall | | | Training data | |
|---|---|---|---|---|---|
| | 3m | 5m | 7m | Occurrences | Percentage |
| *Hardhat* | 98% / 99% | 99% / 93% | 97% / 98% | 3367 | 33% |
| *Safety vest* | 100% / 100% | 96% / 100% | 100% / 100% | 2798 | 27% |
| *Safety gloves* | 100% / 57% | 100% / 41% | 100% / 17% | 2362 | 23% |
| *Safety glasses* | 100% / 18% | 100% / 4% | 100% / 2% | 897 | 9% |
| *Hearing protection* | 100% / 44% | 100% / 9% | 0% / 0% | 860 | 8% |

## 3.3 Experiments and Results

Our primary focus in this paper is detecting personal protective equipment (PPE). Four test phases were made to improve the results iteratively after identifying the drawbacks of the previous phase and making opportune adjustments in the training and/or test data. Each phase employs data captured at different conditions in terms of environment, distance, light, people pose, etc., as reported in Table 1. Different YOLO models have been trained with the indicated type of training data on each phase, reporting the precision and recall in detecting the different target classes (IoU threshold 0.5). The models were trained during 10000 iterations via Google Colab (12-15 hours per model, depending on the dataset size and GPU available). The input image size is 476×476 (RGB images). The training parameters were selected according to (Fernando and Sotheeswaran, 2021): learning rate=0.001, max_batches=10000 (number of iterations), steps=8000, 9000 (iterations at which the learning rate is adjusted), filters=30 (kernels in convolutional layers). Tests, including FPS computation, were done on a laptop with an 11th Gen Intel(R) Core(TM) i7-1185G7 @ 3.00GHz 1.80 GHz CPU, Intel(R) Iris(R) Xe Graphics GPU, 16GB of RAM, and Windows 10 Enterprise.

**First test phase.**

Phases one and two (below) have acted as guidelines to refine the experiments. The training data is from the databases described in Section 3.1. Results are given in Table 2, including the number of training occurrences of each object. The test data consisted of images from three different people (two males, one female) that we acquired with a smartphone, with a camera-person distance of 3, 5, and 7 meters. Some examples are given in Figure 3 (left). Each person wears all PPE from different angles, lighting, and distances, with a total amount of 119 images (3 meters), 127 (5 meters), 111 (7 meters). In general, the images have backgrounds with inferences of different nature (including other people) and poor lighting due to shadows or light sources in the background. With this test data, we aimed at simulating uncontrolled imaging conditions, as those shown in Figure 2, albeit indoor. From the results, we observe a high precision in most classes and distances, but the recall is low for three objects (safety gloves, glasses, and hearing protection) regardless of the distance. High precision but low recall means that the majority of positive classifications are positives, but the model detects only some of the positive samples (in our case, it means missing many PPE occurrences). These three objects do not just appear smaller than hardhats of safety vests in the crop of a person, but they are also under-represented in the training set (at least glasses and hearing protection). Safety vests occupy a significant portion of the body, and they have a significant shape and color. Hardhats may not occupy a big portion, but they have the same dimension ratio and are usually visible from any pose, facilitating their detection at any distance. On the other hand, the bounding box and appearance of safety gloves, glasses, and hearing protection may be different from different viewpoints of the camera, or suffer from occlusions by body parts. As a result, their detection is significantly more difficult.

Table 3: Results of the second test phase. The number of occurrences in brackets indicates the increase with respect to the first test phase.

| Object | Precision/Recall | | | Training data | |
|---|---|---|---|---|---|
| | 3m | 5m | 7m | Occurrences | Percentage |
| Hardhat | 100% / 100% | 99% / 93% | 94% / 95% | 3636 (+269) | 29% |
| Safety vest | 100% / 100% | 98% / 100% | 100% / 100% | 2912 (+114) | 24% |
| Safety gloves | 100% / 59% | 100% / 53% | 100% / 17% | 2891 (+265+264) | 23% |
| Safety glasses | 100% / 18% | 100% / 8% | 0% / 0% | 1418 (+260+261) | 12% |
| Hearing protection | 100% / 55% | 100% / 9% | 0% / 0% | 1482 (+311+311) | 12% |

Table 4: Results of the third test phase.

| Object | 3m | | 5m | | Training data | |
|---|---|---|---|---|---|---|
| | Prec/Rec | AP | Prec/Rec | AP | Occurrences | Percentage |
| Hardhat | 100% / 100% | 100% | 100% / 100% | 100% | 3636 | 29% |
| Safety vest | 100% / 100% | 100% | 100% / 100% | 100% | 2912 | 24% |
| Safety gloves | 100% / 96% | 96% | 100% / 96% | 89% | 2891 | 23% |
| Safety glasses | 100% / 62% | 27% | 100% / 60% | 21% | 1418 | 12% |
| Hearing protection | 100% / 76% | 99% | 100% / 16% | 76% | 1482 | 12% |
| | mAP: | 85% | mAP: | 77% | | |

**Second test phase.**

To analyze the effect of the class imbalance on the performance, this section will consider additional training data via augmentation to include more occurrences of the different objects and increase the proportion of under-represented classes. Some images have been modified in brightness, contrast, and sharpness as follows (affecting mostly the three under-performing classes of the first phase):

- 269 hardhat images had +20% brightness, +10% contrast, and sharpened
- 114 safety vest: +10% brightness, -20% contrast, and blurred
- 265 safety gloves: +40% brightness, +30% contrast, and sharpened
- 264 safety gloves: -20% brightness, -30% contrast, and blurred
- 260 safety glasses: +25% brightness, -25% contrast, and blurred
- 261 safety glasses: -15% brightness, +40% contrast, and sharpened
- 311 hearing protection: +30% brightness, -40% contrast, and blurred
- 311 hearing protection: -30% brightness, +10% contrast, and sharpened

The same test data from phase one is also used here. Results of the second test phase, including the training occurrences after adding the mentioned augmentation, are given in Table 3. The frames per second were measured at 0.7 with the test laptop at this stage of the process.

The two classes that were doing well previously (hardhat and safety vest) remain the same here. Only hardhat does a little worse at 7 meters. Safety gloves have slightly improved recall at 3 and 5 meters, although still 41% and 47% of the positives are undetected. At 7 meters, it keeps without detecting 83% of the occurrences. The improvement of safety glasses is negligible, missing the majority of occurrences as in the previous phase, and at 7 meters, it does not detect any. The small size and transparency of safety glasses may be the reason why they are difficult to detect, which may be affected by variations in reflections under different lighting. Regarding hearing protection, it shows some improvement at 3 meters (recall goes from 44 to 55%), but at 5 and 7 meters, the system is unusable. The shape, size, and visibility of hearing protection can vary from different angles, as mentioned in the previous test phase, making detection more difficult.

**Third test phase.**

Overall, the data augmentation of the previous test phase contributed to slightly improving two difficult objects (safety gloves and hearing protection) at close distances (3 meters). The recall values were 59% and 55%, respectively, meaning that at least, more than half of the occurrences are detected. Safety glasses remain difficult, and in any case, longer distances (5 and especially 7 meters) are 'reserved' for the less difficult hardhat and safety vest classes only.

Table 5: Results of the fourth test phase. The number of occurrences in brackets indicates the increase with respect to the second and third test phases.

| Object | 3m | | 5m | | Training data | |
|---|---|---|---|---|---|---|
| | P/R | AP | P/R | AP | Occurrences | Percentage |
| Hardhat | 100% / 100% | 100% | 100% / 100% | 100% | 3636 | 25% |
| Safety vest | 100% / 100% | 100% | 100% / 100% | 100% | 2912 | 20% |
| Safety gloves | 100% / 100% | 98% | 100% / 99% | 91% | 2891 | 20% |
| Safety glasses | 100% / 96% | 97% | 100% / 58% | 59% | 2523 (+1105) | 17% |
| Hearing protection | 100% / 100% | 99% | 100% / 54% | 94% | 2587 (+1105) | 18% |
| | | mAP: 99% | | mAP: 89% | | |

To isolate the difficulties imposed by operation in uncontrolled environments, we captured new test data in a more controlled environment. Consequently, we gathered additional testing data from two new persons in a controlled environment with a less disturbing background (Figure 3, right), in particular 100 images from 3 meters, and 100 images from 5 meters. A test at 7 meters is not relevant for the intended scenario. The camera is positioned at head-level height, with the person in frontal or nearly frontal view. Both the subject and the camera are static to reduce blurriness. We also used a more uniform white background (a wall) to eliminate inferences. Our aim with this research is to enable a system like the one depicted in Figure 1, where PPE detection and facial recognition can be used with collaborative subjects in an entry access point to a restricted area. This setup is similar e.g. to entry kiosks with face biometrics control, where the person is asked to stand briefly while being pictured, and non-frontal view angle is unusual. Two add-ons here are that *i*) the person can stay at a certain distance of 3 or 5 meters for increased convenience, and *ii*) we add, via our developments in this paper, a PPE compliance check. The training set employed is the same as in the second phase.

Detecting the person and cropping the bounding box before doing PPE detection is also expected to impact accuracy and speed results positively. Thus, from this phase, we added a human detection and cropping step, which was not included in the previous phases. This was achieved via the CVZone library in Python3, which uses OpenCV at the core. Cropping people before detecting any objects reduces the information that must be sent or processed, allowing for higher frame rates or usage of less powerful solutions, such as edge computing. This also improves execution time and detection accuracy of further modules (Nath et al., 2020), e.g., by eliminating the risk of potential background objects similar to PPE, or detecting objects not worn by persons. The algorithm for human detection is based on Histogram of Oriented Gradients (HOG). PPE detection is then achieved with YOLOv4. It can also be combined with face recognition to produce virtual key cards for workers and allow access only if the right person enters with the appropriate gear (Figure 4).

The new test environment of this phase achieved the results shown in Table 4. The system ran with a frame per second rate of 1.2, higher than the 0.7 of the previous phase. We attribute this improvement to adding the human detection and cropping step, which reduces the input information to the YOLO detector. As observed previously, hardhat and safety vest detection is possible with very high accuracy, achieving perfect detection in the controlled environment of this phase. The safety gloves improved the recall from 59% at 3m and 53% at 5m to 96% at both distances, meaning that this item would be missed on a few occasions only in the new scenario. The safety glasses also improved the recall from 18/8% (at 3/5m) to 62/60%, respectively. The recall for hearing protection increased from 55/9% (at 3/5m) to 76/16%, respectively, meaning that detection at a long distance (5 meters) remains difficult.

In this phase, we also added Average Precision (AP) results, which correspond to the area under the precision-recall curve, providing a single metric that does not consider the selection of a particular threshold in the IoU. The Mean Average Precision (mAP) is also given, as the average of APs from all classes. The AP values are high for all classes, except for safety glasses which stay at 27/21% for 3/5m. In comparison to the other objects, safety glasses can be transparent and tiny, making more challenging to find them even if we control the acquisition process. Hearing protection also has a small AP at 5 meters compared to the other objects. As pointed out in the previous phases, this object is smaller than others, which will be further magnified at longer distances. In addition, it can be partially occluded by the face, even with minor unnoticeable departures from frontal viewpoint. Hearing protection is also an underrepresented class in the training set (despite efforts to augment it), which could also be part of the issue.

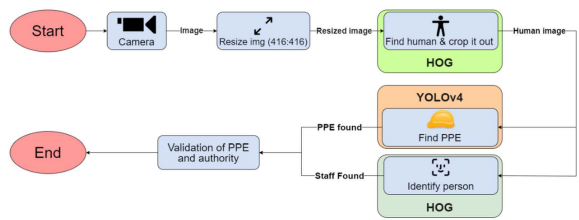

Figure 4: Illustration of the intended pipeline of our system.

**Fourth test phase.**

In this phase, we aim to improve further the representation of classes with fewer images (safety glasses and hearing protection) via additional training data. To do so, new images from two of the contributors of phase one were captured, this time only wearing safety glasses and hearing protection. In particular, 110 images of one person and 111 images of another person were captured. The first 110 images have also been augmented four times as follows:

- +30% brightness, +30% contrast, and sharpened
- -40% brightness, +20% contrast, and sharpened
- +20% brightness, -10% contrast, and sharpened
- +30% brightness, +60% contrast, and blurred

and the other 111 images as:

- -10% brightness, -20% contrast, and blurred
- -40% brightness, and sharpened
- +50% brightness, +50% contrast, and sharpened
- +40% brightness, and blurred

This results in 1105 new occurrences of safety glasses and hearing protection. The new images were taken in a similar test environment as phase three (frontal view, camera at head level, static) to consider our intended conditions of a collaborative setup like Figure 1. The same test data from phase three and person cropping is also used here. The two persons contributing to the training set of this phase are different than the two persons in the test set, ensuring independence. The results obtained with these new training conditions are shown in Table 5. As can be seen, any eventual uneven distribution of object occurrences in the training set is now corrected. The frames per second with the test laptop were measured at 1.2.

The results show that, with the addition of new training images with safety glasses and hearing protection, detection of all classes is feasible at 3 meters (recall of 96% or higher). Hearing protection also increases recall at 5 meters from 16% to 54%, although the system struggles to find safety glasses and hearing protection at this distance, which we still attribute to the factors mentioned earlier: smaller size, transparency (glasses) or occlusion by the head (hearing protection). The new training environment contributed to better AP and mAP values as well, with an mAP of 99% (3 meters) and 89% (5 meters), surpassing by a great margin the results of the previous phase.

## 4 CONCLUSIONS

This work studies the use of cameras to detect Personal Protective Equipment (PPE) usage in industrial environments where compliance is necessary before entering certain areas. For this purpose, we have collected appropriate data, and evaluated several detection models based on the popular YOLOv4 (Bochkovskiy et al., 2020) architecture. We contribute with a study where the number of target classes (hardhat, safety vest, safety gloves, safety glasses, and hearing protection) is higher than in previous works, which usually focus only on detecting one or two (typically hardhats or vests). To aid in the detection accuracy and speed, we also incorporate human detection and cropping, so the PPE detector only has to concentrate on regions with a higher likelihood of a person wearing the target equipment.

We have carried out four test phases of our system, improving the results iteratively after identifying potential drawbacks in the previous one. Test data with people wearing the complete PPE set and not appearing in the training set was used to measure the detection accuracy of our system. Based on the results of our last iteration, a well-functioning system is obtained, with a mean average precision of 99% at a 3 meters distance, and 89% at 5 meters working on controlled setups like an airlock room (Figure 1). The system works best at close distances and in a controlled environment (phases 3 and 4), and it is also benefited from data augmentation through modifications in image brightness, contrast, and sharpness. The latter contributes to enriching our training data and balancing under-represented classes.

At longer distances, uncontrolled environments, or without human detection and cropping, the system operates worse. The difficult imaging conditions and the size of some objects are highlighted as potential reasons. Objects such as hardhats and safety vests are well detected regardless of environment and distance, while hearing protection, safety gloves, and especially glasses are more difficult since they can appear smaller or partially occluded by body or face parts. Regarding detection speed, we achieve frame-per-second (FPS) rates of 1.2 in a laptop. This could be sufficient if the accuracy with just one frame is good since the system can operate as the person approaches. Currently, our method carries out detection on a frame-by-frame basis, but more robustness could be obtained by consolidating the result of several frames, especially to detect the most difficult classes. To help in this mission, we could also incorporate tracking methods to analyze image regions where previously there was a detection with high accumulated confidence. However, this would come at the cost of needing more frames, so more powerful hardware would be needed.

Another improvement, in addition to frame fusion or tracking, is to look for PPE objects in areas where they are expected to appear. For example, hardhats, glasses and hearing protection will likely be in the top part of the body (the head), whereas safety vests must occupy a significant portion in the middle. Skeleton detection (Cao et al., 2019) may also assist in finding where safety gloves may be, as well as refine the other elements.

When deploying a detection system like the one in this paper, one must consider the various ethical questions related to camera-based detection, due to humans appearing in the footage. Whenever a camera is capturing or streaming such type of data to a remote location, privacy, security and GDPR concerns emerge. These concerns would be significantly counteracted via edge computing, with data processed as close as possible to where it is being captured, diminishing transmission of sensitive data to a different location through data networks. Also, necessary frames must be deleted as soon as computations are done. The present system only uses one frame, but even combining several frames with sufficient frame rate would mean that the necessary data to be processed only affects a few milliseconds of footage. Handling the data in this way means that no sensitive data would ever be stored, or transmitted elsewhere.

## ACKNOWLEDGEMENTS

This work has been carried out by Jonathan Karlsson and Fredrik Strand in the context of their Bachelor Thesis at Halmstad University (Computer Science and Engineering), with the support of HMS Networks AB in Halmstad. Authors Bigun, Alonso-Fernandez and Hernandez-Diaz thank the Swedish Research Council (VR) and the Swedish Innovation Agency (VINNOVA) for funding their research.